\documentclass{article}
\usepackage{gropd_preprint,times}
\usepackage{amsmath,amsfonts,bm}

\def\eqref#1{(\ref{#1})}
\def\1{\bm{1}}

\DeclareMathAlphabet{\mathsfit}{\encodingdefault}{\sfdefault}{m}{sl}
\SetMathAlphabet{\mathsfit}{bold}{\encodingdefault}{\sfdefault}{bx}{n}

\newcommand{\E}{\mathbb{E}}

\newcommand{\KL}{D_{\mathrm{KL}}}

\usepackage{amsmath,amssymb}
\usepackage{graphicx}
\usepackage{float}
\usepackage{xcolor}
\usepackage{colortbl}
\usepackage{hyperref}
\usepackage{url}
\usepackage{microtype}
\hypersetup{
  hidelinks,
  pdftitle={GR-OPD: Group-Relative Sequence Credit for On-Policy Distillation},
  pdfauthor={Yushu Wang, Renkai Yang, Jingcao Xu, Zili Geng, Haihong Tang, Yaohui Jin, Hao He},
  pdfsubject={On-policy distillation of language models},
  pdfkeywords={language models, on-policy distillation, sequence credit, teacher feedback}
}
\newtheorem{proposition}{Proposition}

\newcommand{\method}{\textsc{GR-OPD}}

\newcommand{\setIIFour}{\ensuremath{\mathrm{I}_{4}^{\mathrm{nothink}}\!\to\!\mathrm{I}_{1.7}^{\mathrm{nothink}}}}
\newcommand{\setIBFourThink}{\ensuremath{\mathrm{I}_{4}^{\mathrm{think}}\!\to\!\mathrm{B}_{1.7}}}
\newcommand{\setIBThirtyTwoThink}{\ensuremath{\mathrm{I}_{32}^{\mathrm{think}}\!\to\!\mathrm{B}_{1.7}}}
\newcommand{\setIIThirtyTwo}{\ensuremath{\mathrm{I}_{32}^{\mathrm{nothink}}\!\to\!\mathrm{I}_{1.7}^{\mathrm{nothink}}}}
\newcommand{\toprule}{\hline}
\newcommand{\midrule}{\hline}
\newcommand{\bottomrule}{\hline}

\title{GR-OPD: Group-Relative Sequence Credit for On-Policy Distillation}

\newcommand{\gropdauthorblock}[3]{%
  \begin{minipage}[t]{0.25\textwidth}
    \centering
    {\normalsize\bfseries #1$^{#2}$}\\[2pt]
    {\fontsize{8.5}{10}\selectfont\href{mailto:#3}{#3}}
  \end{minipage}%
}

\author{%
\parbox[t]{\textwidth}{\centering
  \gropdauthorblock{Yushu Wang}{1}{booker@sjtu.edu.cn}%
  \gropdauthorblock{Renkai Yang}{2,\dagger}{renkai.yrk@taobao.com}%
  \gropdauthorblock{Jingcao Xu}{2,*}{xujingcao.xjc@taobao.com}%
  \gropdauthorblock{Zili Geng}{2,*}{gengzili.gzl@taobao.com}\\[7pt]
  \gropdauthorblock{Haihong Tang}{2}{piaoxue@taobao.com}%
  \gropdauthorblock{Yaohui Jin}{1}{jinyh@sjtu.edu.cn}%
  \gropdauthorblock{Hao He}{1,\dagger}{hehao@sjtu.edu.cn}\\[7pt]
  {\small
    $^{1}$Shanghai Jiao Tong University, Shanghai, China\\[2pt]
    $^{2}$Alibaba Group, Hangzhou, China}
}}

\makeatletter
\newcommand{\gropdauthornotes}{%
  \begingroup
  \renewcommand{\thefootnote}{}%
  \long\def\@makefntext##1{\noindent##1}%
  \footnotetext{%
    $^{*}$Equal contribution.\\[1pt]
    $^{\dagger}$Corresponding authors.
  }%
  \endgroup
}
\makeatother

\begin{document}
\maketitle
\gropdauthornotes
\raggedbottom

\begin{abstract}
On-policy distillation (OPD) provides token-level teacher feedback, yet common sampled-token updates assign credit locally and omit the effects of earlier actions on subsequent feedback. We introduce \method{} (Group-Relative On-Policy Distillation), which decouples feedback computation from credit assignment by aggregating token feedback into calibrated sequence-level credit. We establish a theoretical basis for this approach by showing that, under ideal on-policy assumptions, broadcasting raw summed feedback to every response token yields an unbiased estimate of the negative gradient of the sequence-level reverse-KL objective. To support effective learning in practice, \method{} discounts later feedback that may become less reliable under distribution shift and uses same-prompt rollout groups to calibrate teacher guidance. It preserves relative response preferences while retaining a controlled fraction of the group mean, which we term collective teacher pressure. Across five Qwen3 teacher--student settings and six mathematical reasoning benchmarks, \method{} outperforms vanilla OPD in macro Avg@K in every setting, including gains of 10.9 and 11.5 percentage points in two settings with matched thinking patterns. It also maintains stable training under mismatched thinking patterns, where vanilla OPD collapses. Ablations show that retaining an appropriate amount of collective teacher pressure can improve learning, while excessive collective teacher pressure or an unrestricted feedback horizon can destabilize training.
\end{abstract}

\section{Introduction}

On-policy distillation (OPD) has become a common method for post-training large language models (LLMs). Recent frontier LLMs including DeepSeek-V4, MiMo-V2-Flash, and Kimi K3 use OPD to transfer or integrate reasoning capabilities~\citep{deepseek_v4,mimo_v2,kimi_k3}. OPD samples responses from the student policy and trains the student to match the teacher's next-token distributions along the generated trajectories. A widely used lightweight variant uses the teacher--student log-probability difference of each student-sampled token as its token-level advantage in a policy-gradient update, avoiding the memory cost of retaining full-vocabulary distributions at every position. Beyond computational efficiency, this sampled-token formulation supports strong post-training gains and provides dense teacher supervision throughout generation, helping to alleviate the credit assignment problem.

Recent work has explored how to make dense teacher supervision more effective in OPD. These efforts include reducing the variance of sampled-token updates~\citep{vopd} and addressing the declining reliability of teacher feedback along student-generated trajectories~\citep{teacher_cannot}. TOPD also addresses limitations of isolated token-level updates by incorporating near-future trajectory guidance~\citep{topd}. Other approaches refine the transferred signal by isolating changes induced by teacher post-training~\citep{opd2_delta} or correcting systematic discrepancies associated with model families and styles~\citep{compassopd,lightning_opd2}. These studies suggest that dense teacher feedback is not uniformly beneficial: useful guidance can coexist with teacher-specific preferences that the student cannot readily absorb. Effective distillation therefore calls for considering both how dense teacher feedback is organized into learning signals and how these signals are adapted to the student's current capabilities and reasoning patterns.

Our key insight is that the granularity of teacher feedback computation and the granularity of credit assignment are separate design choices. OPD can compute teacher feedback for every token, but this does not require assigning each action credit only from the feedback computed for that token. We therefore investigate how to aggregate token-level teacher feedback across a response into a trajectory-level signal and use it to assign credit to the actions that produced that response. This allows later teacher feedback to inform earlier decisions while retaining dense token-level scoring. Evidence supports the informativeness of response-level scores: Rethinking OPD finds that averaging teacher feedback over a response distinguishes correct from incorrect answers even when OPD training fails under mismatched thinking patterns~\citep{rethinking}. We observe the same phenomenon in our teacher--student settings with both matched and mismatched thinking patterns.

To understand how aggregation changes credit assignment, we revisit the gradient of the reverse-KL objective underlying OPD. An early action affects both the immediate teacher--student agreement and the later prefixes visited by the student. Building on MiniLLM and Distributional DAgger~\citep{minillm,distil}, we separate this gradient into local action matching and future state-visitation credit. The common sampled-token update retains only the local term. We show that, under ideal on-policy assumptions, broadcasting the sequence credit obtained by summing teacher feedback to every token yields an unbiased estimate of the negative gradient of OPD's reverse-KL objective. This result establishes the credit-assignment basis for trajectory aggregation and motivates the practical estimator developed below.

Building on this analysis, we propose \method{} (Group-Relative On-Policy Distillation), which converts token-level teacher feedback into sequence-level credit that supports stable and effective learning. Our observation that directly broadcasting uncalibrated trajectory-level signals can lead to training collapse motivates two complementary controls. To reduce the influence of later feedback that may become less reliable under distribution shift, \method{} applies exponential discounting with parameter $\gamma$. To adapt teacher guidance to the student, \method{} uses the same-prompt group mean as a reference for separating relative response preferences from a shared score component. This shared component shifts the credit assigned to all responses in the group together; we refer to it as \emph{collective teacher pressure}. \method{} preserves the relative preference term and adds a fraction $\lambda$ of collective teacher pressure before applying a common group normalization. We evaluate \method{} across five Qwen3 teacher--student settings and six mathematical reasoning benchmarks. \method{} substantially outperforms vanilla OPD in macro Avg@K across all five settings.

Our contributions are threefold:
\begingroup
\setlength{\leftmargini}{1.5em}
\setlength{\topsep}{3pt}
\setlength{\partopsep}{0pt}
\begin{itemize}
\setlength{\itemsep}{6pt}
\setlength{\parsep}{0pt}
\setlength{\parskip}{0pt}
\item \textbf{Recasting OPD through sequence-level credit.} We develop a principled sequence-credit formulation of OPD that transforms dense token-level teacher feedback into a response-level signal. By allowing subsequent feedback to inform earlier actions, this formulation captures trajectory-wide effects beyond local token matching. Under ideal on-policy assumptions, broadcasting summed teacher--student log-ratios recovers the full reverse-KL descent gradient in expectation, grounding sequence-level credit in OPD's original objective.
\item \textbf{Calibrating sequence credit with GR-OPD.} We propose the \method{} estimator to calibrate sequence credit by combining feedback-horizon control with collective teacher pressure calibration. Same-prompt groups provide a reference for relative response preferences and collective teacher pressure. The estimator preserves the relative preference term while regulating how much of the shared component enters the student update.
\item \textbf{Strong long-reasoning performance and stable training.} \method{} outperforms vanilla OPD in macro Avg@K across all five teacher--student settings, including gains of 10.9 and 11.5 percentage points in two settings with matched thinking patterns. It also maintains stable training under mismatched thinking patterns, where vanilla OPD collapses. Ablations show that retaining appropriate collective teacher pressure can improve learning, whereas excessive pressure or a full feedback horizon can destabilize training.
\end{itemize}
\endgroup

\section{Related Work}
\begingroup
\setlength{\parskip}{0pt}

\noindent\textbf{On-policy distillation.} MiniLLM and generalized knowledge distillation (GKD) established student-generated trajectories as an effective basis for language-model distillation~\citep{minillm,gkd}. Subsequent work extends this paradigm to multi-teacher and multimodal learning~\citep{uniopd}, self-distillation with privileged context~\citep{opsd}, and asynchronous training~\citep{fopd}. Alongside these extensions, mechanistic studies investigate gradient variance, teacher--student compatibility, and training pathologies~\citep{vopd,rethinking,demystifying}, clarifying when dense teacher supervision translates into effective learning.

\noindent\textbf{Sequence-level credit assignment.} DAgger motivates learning on states visited by the student~\citep{dagger}. MiniLLM and Distributional DAgger further distinguish local action matching from the influence of actions on future state visitation~\citep{minillm,distil}. To address the limitations of local token-level supervision, TOPD incorporates near-future teacher continuations through optimal-transport alignment to identify trajectory divergence and guide local updates~\citep{topd}. \method{} takes a sequence-level approach to credit assignment, aggregating teacher feedback across the tokens of each student-sampled response so that later feedback can inform earlier actions. Our analysis connects raw-sum broadcasting to the complete sequence-level reverse-KL gradient under ideal on-policy assumptions.

\noindent\textbf{Group-relative optimization and distillation.} GRPO and REINFORCE with a leave-one-out baseline use multiple responses to the same prompt to construct relative learning signals~\citep{deepseekmath,rloo}. Within distillation, GC-OPD separately normalizes trajectory-level OPD scores and verifier rewards within rollout groups, then distributes their disagreement across token-level advantages~\citep{gcopd}. \method{} uses same-prompt groups as a reference for expressing relative response preferences in trajectory-level teacher feedback.

\noindent\textbf{Teacher-signal calibration.} G-OPD reformulates reverse-KL distillation as dense KL-regularized reinforcement learning and tunes the balance between the implicit reward and KL regularization~\citep{g_opd}. Other approaches address the content of teacher guidance: CompassOPD uses a teacher-family reference to remove cross-family offsets~\citep{compassopd}; Lightning OPD 2.0 subtracts recurring style-related disagreement~\citep{lightning_opd2}; and On-Policy Delta Distillation isolates the teacher's log-probability changes from its pre-tuning checkpoint~\citep{opd2_delta}. \method{} calibrates the shared component represented by the same-prompt mean, retaining a controlled fraction $\lambda$ of this collective teacher pressure in the advantages.

\noindent\textbf{Feedback reliability and granularity.} Teacher supervision can become less reliable as student-generated prefixes grow longer, motivating early-stopped rollouts and position-dependent weighting~\citep{esr,teacher_cannot,position_bias}. FiRe-OPD combines trajectory filtering with token-level reweighting~\citep{fire}, while position-weighted self-distillation adjusts supervision according to token position~\citep{pw_opsd}. These studies motivate accounting for both feedback reliability and optimization granularity when converting dense teacher supervision into learning signals for the student's actions throughout a long response.

\endgroup

\section{From Token Credit to Sequence Credit}
\label{sec:local}
\begingroup
\setlength{\parskip}{6pt}

\subsection{Motivation}

Sampled-token OPD assigns each action only its local feedback $r_t$, the teacher--student log-ratio at that token. It therefore cannot use evidence revealed later in the response to revise the credit assigned to an earlier decision. Dense token-level feedback need not be used only for local credit assignment. Sequence aggregation addresses this limitation by pooling teacher feedback over the trajectory and broadcasting the resulting signal to the actions that produced it. Figure~\ref{fig:sequence-credit-training}(a) illustrates the distinction: local feedback favors the incorrect first action $B$, whereas aggregation reverses the preference once later evidence is included, assigning greater credit to the action that leads to the correct response $A$.

This distinction is particularly relevant to cross-pattern distillation, where different reasoning patterns can make the teacher's prefix-level preferences poorly aligned with the student's actions. Sequence aggregation does not remove the underlying pattern mismatch; it integrates token-level teacher feedback distributed across the response, allowing the sequence-level preference to counteract isolated, locally misleading signals. Rethinking OPD provides empirical support for this view: sequence-mean teacher feedback distinguishes correct from incorrect student responses with an AUROC of $.75$ under a cross-pattern setting, comparable to $.73$ under a setting with matched thinking patterns~\citep{rethinking}. We also observe this phenomenon in our settings. Across all five teacher--student settings, including both with mismatched thinking patterns, Figure~\ref{fig:sequence-credit-training}(b) reports $J_H(H)$, which measures how strongly teacher scores from the first $H$ tokens favor correct responses within a prompt. Higher values indicate stronger separation; Appendix~\ref{app:fixed-response-diagnostic} gives the definition.

\begin{figure}[t]
\centering
\includegraphics[width=\textwidth,trim=0 0 0 2.5pt,clip]{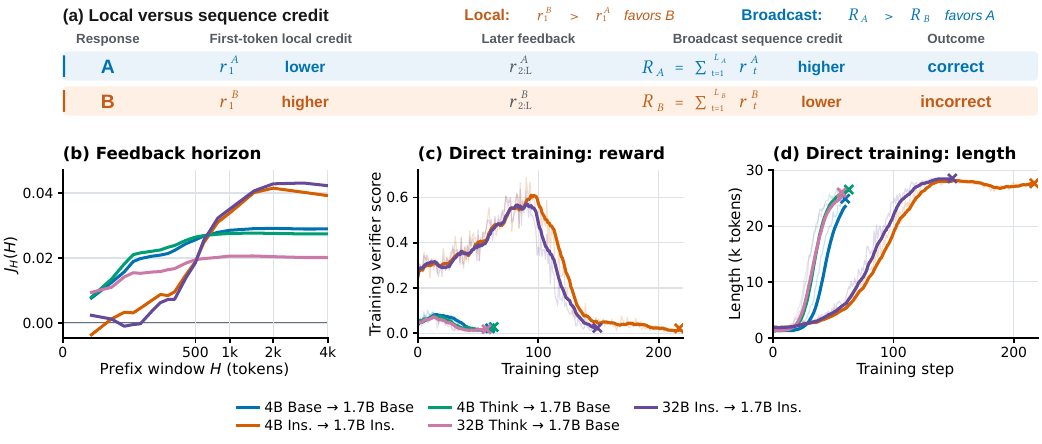}
\caption{Sequence credit and training stability across five teacher--student pairings. (a) Illustrative local feedback favors incorrect $B$, whereas trajectory credit favors correct $A$; outcomes do not enter the objective. (b) $J_H(H)$ measures how strongly prefix scores favor correct responses (higher is better). On fixed responses, it peaks around 1,500--3,000 tokens; the horizontal scale is cube-root. (c,d) Training directly on the discounted trajectory score causes verifier reward to fall as response length approaches the generation cap.}
\label{fig:sequence-credit-training}
\end{figure}

These observations support the informativeness of aggregated teacher feedback. We next develop a theoretical analysis of how trajectory aggregation affects gradient estimation for the reverse-KL objective underlying OPD.

\setlength{\parskip}{3pt}
\subsection{Unbiased Trajectory Broadcast}

For prompt $x$, let $p_\theta$ be the student, $q$ a fixed teacher, and $y=(y_1,\ldots,y_L)$ a student-sampled response with histories $h_t=(x,y_{<t})$. Define token feedback $r_t=\log q(y_t\mid h_t)-\log p_\theta(y_t\mid h_t)$. We revisit the reverse-KL objective underlying OPD:
\begin{equation}
\mathcal K(\theta)=\KL\!\left(p_\theta(\cdot\mid x)\,\|\,q(\cdot\mid x)\right).
\label{eq:seq-kl}
\end{equation}
Let $z_t=\nabla_\theta\log p_\theta(y_t\mid h_t)$ denote the parameter gradient of the sampled token's log-probability; multiplying $z_t$ by its assigned credit gives that token's contribution to the policy update. All expectations below are over responses sampled from $p_\theta(\cdot\mid x)$.

\noindent\textbf{Local feedback captures only part of sequence credit.} An action changes both the teacher--student agreement at its own position and the later prefixes visited by the student. Building on MiniLLM and Distributional DAgger~\citep{minillm,distil}, we write
\begin{equation}
-\nabla_\theta\mathcal K
=\underbrace{\E\!\left[\sum_t z_t r_t\right]}_{\text{local action matching}}
+\underbrace{\E\!\left[\sum_t z_t\sum_{u>t}r_u\right]}_{\text{future state-visitation credit}}.
\label{eq:sampled-credit}
\end{equation}
The derivation is provided in Appendix~\ref{app:sequence-gradient}. The common sampled-token update retains only the first term. The second attributes later teacher feedback to earlier actions that shaped the subsequent trajectory. We therefore consider broadcasting the full trajectory sum to recover both the local and future components of the reverse-KL gradient.

\begin{proposition}[Unbiased trajectory broadcast]
\label{prop:broadcast}
Under on-policy sampling, broadcasting the summed feedback $R=\sum_t r_t$ to every response token yields an unbiased estimate of $-\nabla_\theta\mathcal K$.
\end{proposition}

Autoregressive factorization gives $\log p_\theta(y\mid x)=\sum_t\log p_\theta(y_t\mid h_t)$, and likewise for $q$. Thus, $R=\log q(y\mid x)-\log p_\theta(y\mid x)$, so $\mathcal K=-\E_{p_\theta}[R]$ by the definition of reverse KL.

\begingroup
\setlength{\abovedisplayskip}{3pt}
\setlength{\belowdisplayskip}{3pt}
\setlength{\abovedisplayshortskip}{0pt}
\setlength{\belowdisplayshortskip}{3pt}
Since the teacher is fixed, $\nabla_\theta R=-\sum_t z_t$. Differentiating the expectation accounts for both the sampling distribution and the dependence of $R$ on $\theta$:
\begin{equation}
\begin{aligned}
-\nabla_\theta\mathcal K
&=\nabla_\theta\E_{p_\theta}[R]
=\E_{p_\theta}\!\left[R\sum_t z_t\right]
+\E_{p_\theta}[\nabla_\theta R]\\
&=\E_{p_\theta}\!\left[R\sum_t z_t\right]
-\underbrace{\E_{p_\theta}\!\left[\sum_t z_t\right]}_{=\,0}
=\E_{p_\theta}\!\left[R\sum_t z_t\right].
\end{aligned}
\label{eq:broadcast}
\end{equation}
\endgroup
The marked expectation equals the gradient of the total probability over all student responses. This total is one for every $\theta$, so its gradient is zero. Moreover, the past feedback included in $R$ contributes zero in expectation when multiplied by the current token's score gradient: it is fixed by the current history, and the conditional mean of that gradient is zero (Appendix~\ref{app:past-feedback}). Thus, broadcasting the same sequence credit $R$ preserves the complete reverse-KL gradient in expectation.
\endgroup

\section{Constructing Reliable Group-Relative Sequence Credit}
\label{sec:gropd}

The previous section shows that broadcasting summed teacher feedback preserves the complete reverse-KL gradient under ideal on-policy assumptions. In practice, however, directly using aggregated feedback for policy updates can still lead to unstable training. This motivates \method{}, an estimator that calibrates trajectory feedback to support stable and effective learning.

\subsection{From Sequence Scores to Stable Learning Signals}
\begingroup
\setlength{\parskip}{4pt}
\setlength{\abovedisplayskip}{6pt plus 1pt minus 1pt}
\setlength{\belowdisplayskip}{6pt plus 1pt minus 1pt}
\setlength{\abovedisplayshortskip}{3pt plus 1pt minus 1pt}
\setlength{\belowdisplayshortskip}{4pt plus 1pt minus 1pt}
% Keep run-in headings and their page-break protection with compact spacing.
\makeatletter
\renewcommand{\paragraph}{\@startsection{paragraph}{4}{\z@}
  {4pt plus 1pt minus 1pt}{-1em}{\normalfont\normalsize\bfseries}}
\makeatother

\paragraph{Teacher feedback can become less reliable along the trajectory.}
Sequence credit incorporates later teacher feedback, which may become less reliable under distribution shift as student-generated prefixes grow longer. Prior work reports deteriorating supervision on later student prefixes and improved stability from truncating rollouts~\citep{esr}; the first $30\%$ of tokens can match full OPD while the last $30\%$ alone learns little~\citep{position_bias}; and teacher fidelity decays deeper in the sequence~\citep{teacher_cannot}.

We next examine how this reliability concern affects aggregated teacher feedback. We rescore the same responses using progressively longer prefixes and measure the resulting scores' within-prompt association with answer correctness. Figure~\ref{fig:sequence-credit-training}(b) shows that this association initially strengthens, then begins to decline beyond pairing-specific peaks between roughly 1,500 and 3,000 tokens. These observations motivate retaining downstream evidence while limiting the influence of later feedback. \method{} therefore aggregates feedback using discounted weights, retaining feedback from the full response while progressively reducing the influence of later positions.

\paragraph{Informative trajectory scores do not guarantee stable learning.}
Reducing the influence of later feedback does not by itself make a trajectory score suitable for direct policy optimization. When we train directly on the discounted trajectory score, all five teacher--student settings exhibit the same failure pattern: the training verifier score rises briefly and then declines, while mean response length approaches the generation cap (Figure~\ref{fig:sequence-credit-training}(c,d)). These results motivate further calibration of trajectory scores before using them as learning signals.

This need for calibration is consistent with recent OPD methods that refine teacher guidance by isolating changes induced by teacher post-training~\citep{opd2_delta} or correcting systematic discrepancies associated with model families and styles~\citep{compassopd,lightning_opd2}. The teacher--student log-ratio can reflect differences in capability, style, and reasoning pattern, which need not all provide equally useful guidance to the student.

We use responses to the same prompt as a reference for calibrating this guidance. Subtracting the group mean from each trajectory score expresses the relative preference among responses, while the group mean represents a shared score component. We refer to this shared component as \emph{collective teacher pressure}. It may contain useful guidance as well as differences incompatible with the student's current behavior, so its magnitude alone does not determine its teaching value. \method{} preserves relative response preferences and adds back a controlled amount of collective teacher pressure to construct the advantages.

\clearpage
\begin{figure}[t]
\centering
\includegraphics[width=\linewidth,trim=0 0 0 6pt,clip]{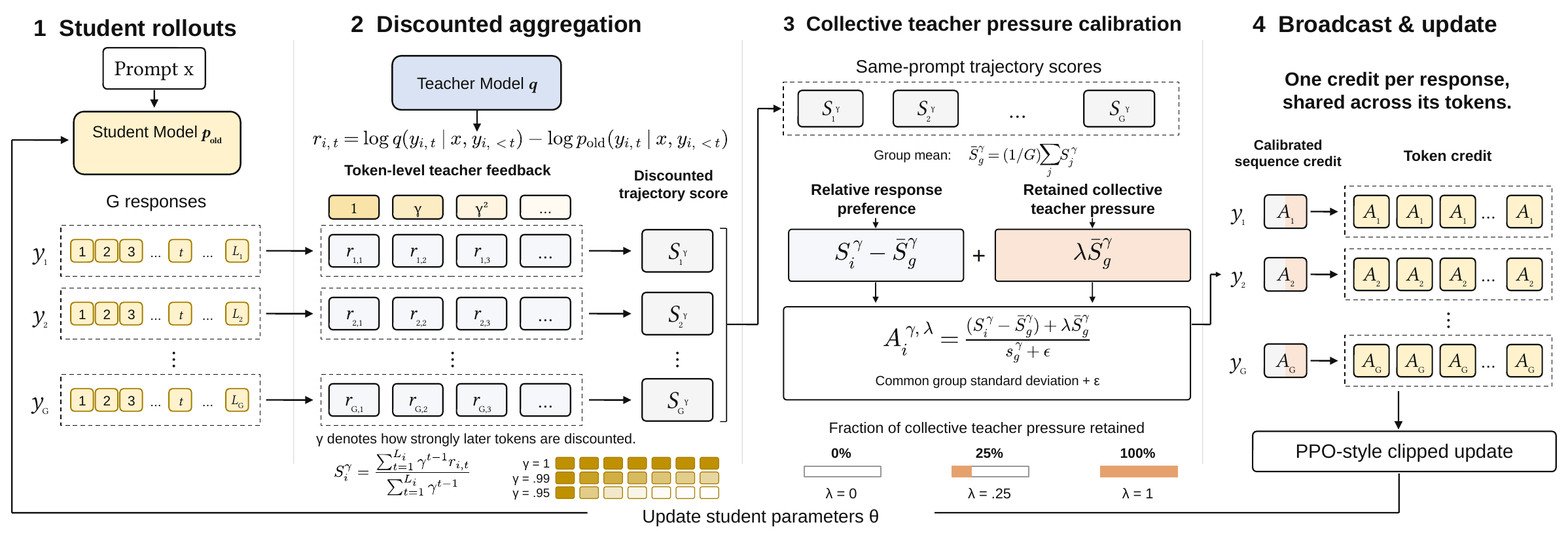}
\caption{\textbf{GR-OPD overview.} (1) The student samples responses to the same prompt and receives token feedback from teacher--student log-probability differences. (2) Discounting with $\gamma$ aggregates feedback into a trajectory score while down-weighting later positions. (3) Calibration preserves relative response preferences and retains a fraction $\lambda$ of the group mean, termed \emph{collective teacher pressure}. The two components are added and normalized by a common group standard deviation. (4) Each response's advantage is broadcast to its tokens for a PPO-style clipped update.}
\label{fig:gropd-method-overview}
\end{figure}

\subsection{The GR-OPD Estimator}
As shown in Figure~\ref{fig:gropd-method-overview}, \method{} comprises four components: student rollouts and teacher feedback, discounted trajectory scoring, group-based calibration, and trajectory broadcast for the policy update.

\paragraph{Student rollouts and teacher feedback.}
For prompt $g$, the rollout policy $p_{\mathrm{old}}$ samples $G$ responses $y_1,\ldots,y_G$. Response $i$ has length $L_i$ and histories $h_{i,t}=(g,y_{i,<t})$. We compute token feedback as $r_{i,t}=\log q(y_{i,t}\mid h_{i,t})-\log p_{\mathrm{old}}(y_{i,t}\mid h_{i,t})$ on the same history.

\paragraph{Discounted trajectory scoring.}
We aggregate token feedback into one score per response:
\begin{equation}
S_i^\gamma
=
\frac{\sum_{t=1}^{L_i}\gamma^{t-1}r_{i,t}}
{\sum_{t=1}^{L_i}\gamma^{t-1}}.
\label{eq:discounted-trajectory-score}
\end{equation}
Dividing by the accumulated discount weight prevents the score from growing merely with response length. Setting $\gamma=1$ gives the full-response mean, whereas $0<\gamma<1$ reduces the influence of later feedback (Appendix~\ref{app:tail-influence}). We empirically choose $\gamma=0.99$ as the default in our experiments.

\paragraph{Group-based calibration.}
Let $\bar S_g^\gamma$ and $s_g^\gamma$ denote the mean and population standard deviation of the $G$ trajectory scores for prompt $g$. For a retention fraction $0\leq\lambda\leq1$, the response-level advantage is
\begin{equation}
A_i^{\gamma,\lambda}
=
\frac{
\underbrace{S_i^\gamma-\bar S_g^\gamma}_{\text{relative response preference}}
+\underbrace{\lambda\bar S_g^\gamma}_{\text{retained collective teacher pressure}}
}
{s_g^\gamma+\epsilon},
\label{eq:gropd-estimator}
\end{equation}
Both terms derive from teacher--student feedback: the first expresses relative response preference, while the second shifts all advantages together and can be positive or negative. The common denominator follows GRPO's within-prompt scaling~\citep{deepseekmath}, with $\epsilon>0$ for numerical stability. At $\lambda=0$, \method{} gives a GRPO-style relative update that favors responses with above-average teacher--student trajectory scores. At $\lambda=1$, the full trajectory score is used with the same scaling, retaining all collective teacher pressure.

\paragraph{Trajectory broadcast and policy update.}
We broadcast $A_i^{\gamma,\lambda}$ to every token in $y_i$, allowing trajectory feedback to inform each action. Feedback and advantages are held fixed during the update. With token probability ratio $\rho_{i,t}(\theta)=p_\theta(y_{i,t}\mid h_{i,t})/p_{\mathrm{old}}(y_{i,t}\mid h_{i,t})$, the PPO-style clipped surrogate~\citep{ppo} is
\begin{equation}
\ell_{i,t}(\theta)
=\min\!\left(
\rho_{i,t}(\theta)A_i^{\gamma,\lambda},
\operatorname{clip}\!\left(\rho_{i,t}(\theta),\,1-\delta,\,1+\delta\right)A_i^{\gamma,\lambda}
\right),
\label{eq:gropd-policy-objective}
\end{equation}
where $\delta$ is the clipping threshold. We maximize the surrogate averaged within each response and then across responses. Thus, $\gamma$ controls the feedback horizon, while $\lambda$ controls the fraction of collective teacher pressure retained.

\endgroup

\clearpage
\raggedbottom
\section{Experiments}
\label{sec:experiments}

\subsection{Setup}
\begingroup
\setlength{\parskip}{0pt}

\noindent\textbf{Teacher--student pairings.} Three pairings match the teacher and student families: Qwen3-4B-Base to Qwen3-1.7B-Base, and Qwen3-4B- or 32B-Instruct (nothink) to Qwen3-1.7B-Instruct (nothink). Two further pairings combine a think-mode Instruct teacher with the Qwen3-1.7B-Base student; we regard these as cross-pattern pairings because the teacher uses Instruct thinking mode while the Base student follows its default reasoning protocol. For compactness, $\mathrm{B}_s$ and $\mathrm{I}_s$ denote Base and Instruct models with $s$ billion parameters, and a superscript records the think or nothink mode of each Instruct model; both Instruct students are used in nothink mode.

\noindent\textbf{Datasets and metrics.} All runs are trained on the DAPO-Math prompt set \citep{dapo}. We evaluate performance on six mathematical reasoning benchmarks: AIME 2024, AIME 2025, AMC 2023, MATH-500 \citep{math}, Minerva Math \citep{minerva}, and OlympiadBench \citep{olympiadbench}, which comprise 1,042 problems in total. We report the macro average of per-problem sample accuracy (Avg@K) across the six suites. We tag a run ``Collapse'' when its response length grows toward the training limit while the task-verifier reward falls to a near-zero level. The tag marks a failed training trajectory.

\noindent\textbf{Implementation details.} Training uses 221 steps and a maximum response length of 28,672 tokens, and evaluation uses the same response length. The main-table \method{} runs use $\gamma=.99$; they retain no collective teacher pressure ($\lambda=0$) for the base-to-base and cross-pattern pairings and retain $25\%$ ($\lambda=.25$) for the matched Instruct pairings. The vanilla sampled-token OPD reference uses the same prompts, teacher, rollout batch, step count, and clipped-objective optimization, and differs only in the credit rule: its per-token advantage is the raw token-level teacher feedback $r_{i,t}$ of Section~\ref{sec:local}, with neither trajectory aggregation nor group normalization. Appendix~\ref{app:details} clarifies pairing modes and reports model-level reference evaluations and feedback-horizon diagnostics.

\endgroup

\subsection{Main results}
\begingroup
\setlength{\parskip}{0pt}

Table~\ref{tab:main} reports the per-suite and aggregate results for the five evaluated pairings, comparing vanilla sampled-token OPD against \method{}.

\noindent\textbf{Broad gains over vanilla OPD.} Across all five evaluated pairings, \method{} raises the macro average and achieves the higher displayed score on every suite. On the compatible pairings, macro Avg@K increases from 23.333 to 27.800 for Base-to-Base, from 34.118 to 45.066 for Qwen3-4B-Instruct, and from 30.320 to 41.774 for Qwen3-32B-Instruct.

\noindent\textbf{Stable learning under pattern mismatch.} In both think-Instruct-to-Base pairings, vanilla OPD collapses, ending at 8.45 and 10.07 macro Avg@K. \method{} remains trainable and reaches 29.571 and 24.463 under the same two pairings.

\noindent\textbf{Larger teachers do not guarantee larger gains.} Holding the student family, teacher mode, $\lambda$, and training setup fixed, scaling the teacher from 4B to 32B lowers macro Avg@K from 29.571 to 24.463 for think-Instruct-to-Base and from 45.066 to 41.774 for nothink-Instruct-to-Instruct. This non-monotonic pattern also appears in vanilla OPD~\citep{rethinking}, indicating that teacher size is an unreliable proxy for teaching value.

\endgroup

\begingroup
\setlength{\intextsep}{6pt}
\begin{table}[H]
\centering
\fontsize{8}{8.5}\selectfont
\setlength{\tabcolsep}{3.5pt}
\caption{Per-suite Avg@K (\%) across five teacher--student pairings. Bold marks the higher score within each pairing; Macro is the unweighted mean over the six suites before rounding. ``Collapse'' marks a failed training trajectory.}
\label{tab:main}
\begin{tabular*}{\linewidth}{@{}l@{\extracolsep{\fill}}rrrrrrr@{}}
\toprule
Method & AIME24 & AIME25 & AMC23 & MATH500 & Minerva & Olympiad & Macro \\
\midrule
\multicolumn{8}{@{}l@{}}{\cellcolor{gray!15}\textbf{\boldmath Qwen3-4B-Base $\rightarrow$ Qwen3-1.7B-Base}} \\
Vanilla OPD & 3.750 & 3.229 & 29.219 & 56.225 & 20.956 & 26.618 & 23.333 \\
\method{} ($\lambda=0$) & \textbf{5.521} & \textbf{3.958} & \textbf{36.094} & \textbf{64.125} & \textbf{26.517} & \textbf{30.588} & \textbf{27.800} \\
\midrule
\multicolumn{8}{@{}l@{}}{\cellcolor{gray!15}\textbf{\boldmath Qwen3-4B-Instruct (nothink) $\rightarrow$ Qwen3-1.7B-Instruct (nothink)}} \\
Vanilla OPD & 17.604 & 12.708 & 44.453 & 68.775 & 20.358 & 40.809 & 34.118 \\
\method{} ($\lambda=.25$) & \textbf{25.000} & \textbf{26.563} & \textbf{58.516} & \textbf{77.550} & \textbf{25.781} & \textbf{56.985} & \textbf{45.066} \\
\midrule
\multicolumn{8}{@{}l@{}}{\cellcolor{gray!15}\textbf{\boldmath Qwen3-4B-Instruct (think) $\rightarrow$ Qwen3-1.7B-Base}\quad\textit{(pattern mismatch)}} \\
Vanilla OPD\,(\textcolor{red!75!black}{Collapse}) & 0.938 & 1.354 & 7.813 & 22.750 & 8.502 & 9.338 & 8.449 \\
\method{} ($\lambda=0$) & \textbf{8.229} & \textbf{5.833} & \textbf{37.891} & \textbf{66.375} & \textbf{25.276} & \textbf{33.824} & \textbf{29.571} \\
\midrule
\multicolumn{8}{@{}l@{}}{\cellcolor{gray!15}\textbf{\boldmath Qwen3-32B-Instruct (think) $\rightarrow$ Qwen3-1.7B-Base}\quad\textit{(pattern mismatch)}} \\
Vanilla OPD\,(\textcolor{red!75!black}{Collapse}) & 1.875 & 3.333 & 13.828 & 23.050 & 6.250 & 12.059 & 10.066 \\
\method{} ($\lambda=0$) & \textbf{4.896} & 3.333 & \textbf{32.266} & \textbf{57.350} & \textbf{21.140} & \textbf{27.794} & \textbf{24.463} \\
\midrule
\multicolumn{8}{@{}l@{}}{\cellcolor{gray!15}\textbf{\boldmath Qwen3-32B-Instruct (nothink) $\rightarrow$ Qwen3-1.7B-Instruct (nothink)}} \\
Vanilla OPD & 12.292 & 11.667 & 40.313 & 63.575 & 18.704 & 35.368 & 30.320 \\
\method{} ($\lambda=.25$) & \textbf{22.188} & \textbf{25.000} & \textbf{57.031} & \textbf{72.850} & \textbf{20.267} & \textbf{53.309} & \textbf{41.774} \\
\bottomrule
\end{tabular*}
\end{table}
\endgroup

\begingroup
\setlength{\intextsep}{3pt}
\begin{figure}[H]
\centering
\includegraphics[width=\linewidth]{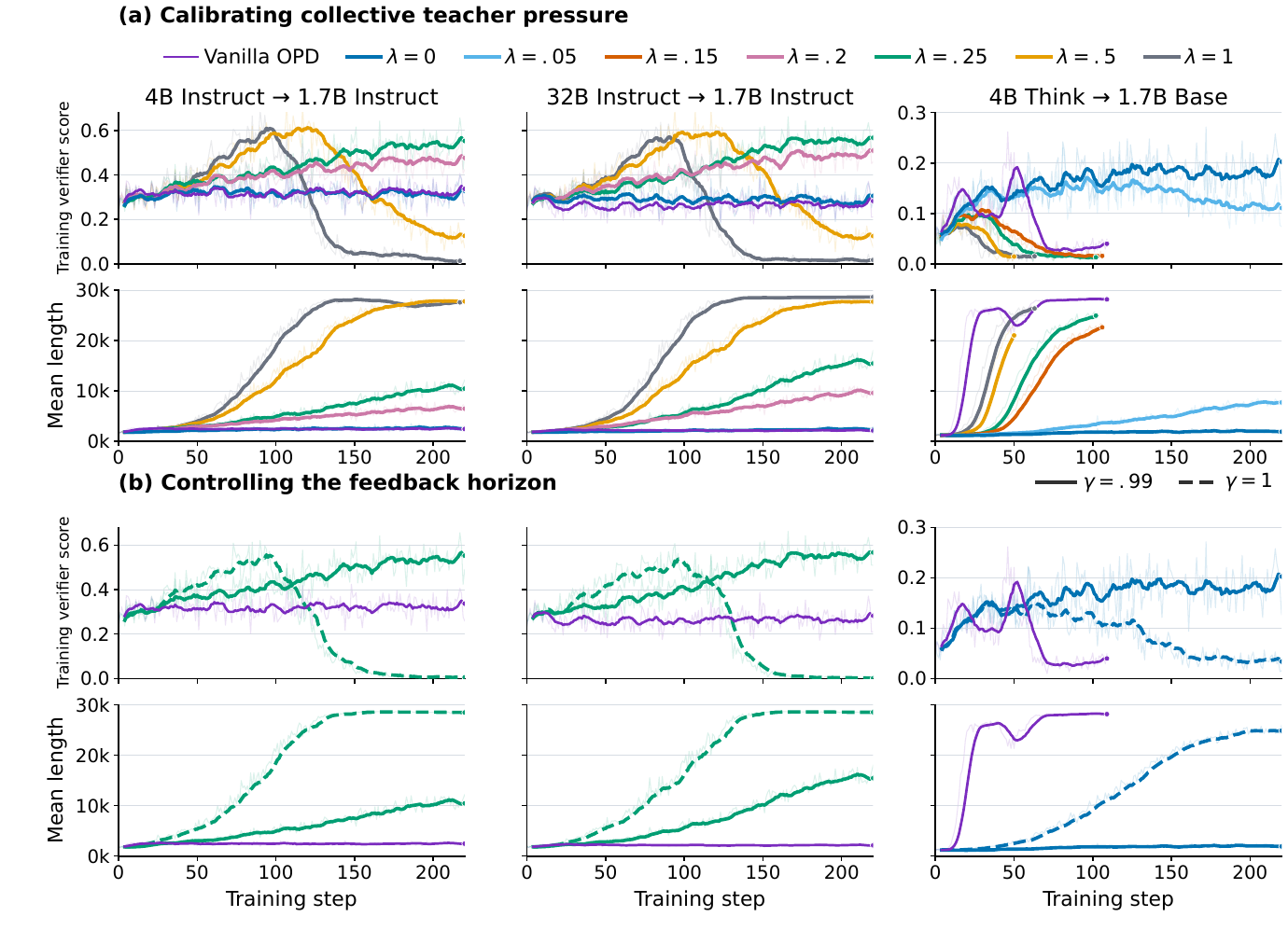}
\caption{Training dynamics across two matched Instruct pairings and one cross-pattern pairing. Purple curves denote vanilla OPD, and GR-OPD colors denote $\lambda$. (a) Calibrating collective teacher pressure at $\gamma=.99$: $\lambda$ is the retained fraction of the shared component. (b) Controlling the feedback horizon: solid and dashed GR-OPD curves denote $\gamma=.99$ and $\gamma=1$. Faint lines show per-step values, bold lines show ten-step moving averages, and endpoint dots mark the last recorded step.}
\label{fig:training-controls-matrix}
\end{figure}
\endgroup

\subsection{Calibrating collective teacher pressure}
\begingroup
\setlength{\parskip}{0pt}

\noindent\textbf{An appropriate level of collective teacher pressure sustains improvement.} In both matched Instruct pairings, removing all collective teacher pressure ($\lambda=0$) keeps response length stable but yields limited gains in training-verifier reward. At $\lambda=.25$ or $.2$, retaining $25\%$ or $20\%$ of collective teacher pressure alongside the relative preference term supports sustained improvements in verifier reward as responses grow longer (Figure~\ref{fig:training-controls-matrix}(a)). Compared with $\lambda=.2$, $\lambda=.25$ generally produces longer responses and a higher final training-verifier reward. The normalized collective teacher pressure estimates in Table~\ref{tab:shared-score-component} are roughly twice as large for the completed cross-pattern runs as for the matched runs; at $\lambda=0$, no collective teacher pressure is retained in their advantages, and training remains stable in both verifier reward and response length.

\noindent\textbf{Excessive collective teacher pressure can cause training collapse.} With full pressure retention ($\lambda=1$), the shared component is fully retained alongside the relative preference term, and both pattern-matched pairings show rapid length growth followed by reward collapse. In the \setIIFour{} pairing, reducing $\lambda$ to $.5$ makes the deterioration less abrupt, but responses still grow excessively long and reward ultimately falls. Even at its training-reward peak, the $\lambda=.5$ checkpoint performs worse in evaluation than the final $\lambda=.25$ model. An early rise in training reward therefore does not establish that the student can sustain the additional collective teacher pressure. One possible explanation is that the student extends its reasoning trajectories with more teacher-preferred content. At an appropriate level of collective teacher pressure, this expansion may produce useful additional reasoning; beyond that range, greater length no longer brings corresponding gains in correctness.

\noindent\textbf{Pattern mismatch requires stronger suppression of collective teacher pressure.} For the mismatched \setIBFourThink{} pairing, even $\lambda=.15$ leads to reward decline and rapid length growth. Reducing $\lambda$ further to $.05$ improves training stability, but the final training-verifier reward remains lower than when collective teacher pressure is fully removed ($\lambda=0$). One possible explanation is that, in such pairings, the shared component carries incompatible reasoning-pattern differences. The result indicates that the amount of collective teacher pressure a student can absorb depends on teacher--student compatibility rather than on a universal retention fraction.

\endgroup

\begingroup
\setlength{\intextsep}{3pt}
\setlength{\abovecaptionskip}{0pt}
\setlength{\belowcaptionskip}{2pt}
\begin{table}[H]
\small
\setlength{\parskip}{0pt}
\caption{Collective teacher pressure before and after calibration ($\gamma=.99$). Cross-pattern runs show larger normalized pressure than matched runs; $\lambda$ controls the amount retained alongside relative response preferences. Both columns report batch-level magnitude proxies (Appendix~\ref{app:pressure-diagnostic}).}
\label{tab:shared-score-component}
\centering
\fontsize{9}{10}\selectfont
\setlength{\tabcolsep}{5pt}
\renewcommand{\arraystretch}{1}
\begin{tabular*}{\linewidth}{@{}l@{\extracolsep{\fill}}ccc@{}}
\toprule
Teacher $\rightarrow$ student & $\lambda$ & Normalized pressure magnitude proxy & Retained pressure magnitude proxy \\
\midrule
\setIIFour & $.5$ & $3.21$ & $1.60$ \\
\setIIFour & $.25$ & $3.16$ & $0.79$ \\
\setIIFour & $.2$ & $3.17$ & $0.63$ \\
\setIIThirtyTwo & $.25$ & $3.61$ & $0.90$ \\
\setIIThirtyTwo & $.2$ & $3.56$ & $0.71$ \\
\setIBFourThink & $0$ & $7.23$ & $0$ \\
\setIBThirtyTwoThink & $0$ & $6.08$ & $0$ \\
\bottomrule
\end{tabular*}
\end{table}
\endgroup

\raggedbottom
\vspace{-6pt}
\subsection{Controlling the feedback horizon}
\begingroup
\setlength{\parskip}{0pt}

\noindent\textbf{A full feedback horizon can destabilize training.} Figure~\ref{fig:training-controls-matrix}(b) compares $\gamma=.99$ and $\gamma=1$ while holding $\lambda$ fixed within each pairing. In the two pattern-matched pairings at $\lambda=.25$, the full-horizon runs initially gain verifier reward, but reward later declines as response length continues to grow. With $\gamma=.99$, reward continues to improve and length grows more gradually. Our appendix analysis (Figure~\ref{fig:horizon-training-dynamics}) shows a corresponding change in the \setIIThirtyTwo{} pairing: under $\gamma=1$, trajectory scores become more associated with response length and less aligned with verifier reward.

\noindent\textbf{Including more late feedback may create a length-based score shortcut.} Length alone cannot raise a trajectory score that averages token feedback. The risk is that $\gamma=1$ gives late feedback the same positional weight as early feedback, even as student-generated prefixes move into states where the teacher may be less reliable. On these prefixes, repetitive or circuitous continuations may receive teacher--student log-ratios above the preceding trajectory average. Adding such continuations raises the trajectory score without advancing the solution. Broadcasting the resulting score across the trajectory can then reinforce the behavior that produced those continuations.

\noindent\textbf{Discounting weakens this feedback loop.} With $\gamma=.99$, later log-ratios have less influence on the trajectory score, while earlier downstream evidence remains available for credit assignment. The pattern-mismatched \setIBFourThink{} pairing makes the distinction clear: both runs use $\lambda=0$, so collective teacher pressure is removed, yet $\gamma=1$ still leads to reward decline and length growth toward the training limit. The $\gamma=.99$ run continues to improve with relatively stable length. Horizon control therefore remains important when the update removes all collective teacher pressure and uses only within-prompt relative advantages: those advantages still depend on which parts of the trajectory supply their evidence.

\endgroup

\section{Conclusion}
We presented \method{}, a sequence-credit estimator for on-policy distillation that decouples token-level feedback computation from credit assignment. We showed that, under ideal on-policy assumptions, broadcasting summed teacher feedback recovers the negative gradient of the sequence-level reverse-KL objective in expectation. \method{} discounts later feedback and calibrates sequence credit by preserving relative response preferences while controlling the retained collective teacher pressure. Experiments across five Qwen3 teacher--student settings demonstrate consistent improvements over vanilla OPD on mathematical reasoning benchmarks and stable training under mismatched thinking patterns.

Our evaluation is limited to mathematical reasoning and Qwen3 models, leaving generalization to broader domains and model families to be established. Our qualitative analysis suggests that the appropriate level of retained collective teacher pressure depends on teacher--student compatibility. Building on these observations, future work could develop principled criteria for selecting $\lambda$ across teacher--student pairings. Moreover, broadcasting a single sequence-level credit does not explicitly differentiate credit among tokens within a response. Future work could combine calibrated sequence-level credit with token-level credit assignment to preserve \mbox{trajectory-wide} information while enabling more localized updates. \method{}'s same-prompt group structure also provides a natural basis for integration with GRPO: shared rollout groups could support both teacher-derived credit and verifier-based rewards, enabling joint optimization of knowledge transfer and task performance.

\section*{AI Use Statement}
Large language models were used to help find and check relevant literature, discuss experimental design and interpret results, organize experimental records, and draft and edit parts of the manuscript. They also assisted with checking mathematical exposition and preparing figures and code. The authors reviewed the AI-assisted work, checked citations and reported results against their sources, and take responsibility for the final manuscript.

\bibliography{references}
\bibliographystyle{iclr2027_conference}

\appendix
\section{Proofs}
\label{app:proofs}

\subsection{Sequence-gradient decomposition}
\label{app:sequence-gradient}

MiniLLM gives a detailed policy-gradient derivation of return-to-go for sequence reverse KL and then separates the single-step and long-term terms~\citep[Sec.~2.2 and Apps.~A.2--A.3]{minillm}. Distributional DAgger independently formulates the corresponding local/future-credit decomposition for on-policy self-distillation~\citep[Sec.~3.2]{distil}. We give the conditional-KL derivation of Eq.~\eqref{eq:sampled-credit} in our notation.

Let $d_\theta(h)=\KL(p_\theta(\cdot\mid h)\|q(\cdot\mid h))$, and write $D_{p,t}(h)$ for the probability of visiting history $h$ before position $t$. Then
\begin{equation}
\mathcal K(\theta)=\sum_t\sum_hD_{p,t}(h)d_\theta(h).
\end{equation}
Differentiating the conditional mismatch contributes the local term $\E\sum_t\nabla_\theta d_\theta(h_t)$. For the state distribution,
\begin{equation}
\nabla D_{p,t}(h)
=D_{p,t}(h)\sum_{u<t}\nabla\log p_\theta(y_u\mid h_u).
\end{equation}
Substitution and exchanging the order of $u$ and $t$ yields
\begin{equation}
\sum_t\sum_h\nabla D_{p,t}(h)d_\theta(h)
=\E\sum_u z_u\sum_{t>u}d_\theta(h_t),
\end{equation}
which is the future state-visitation term.

At a fixed history,
\begin{align}
\nabla d_\theta(h)
&=\nabla\sum_a p_\theta(a\mid h)
\bigl(\log p_\theta(a\mid h)-\log q(a\mid h)\bigr)\\
&=\E_{a\sim p_\theta}\left[
\nabla\log p_\theta(a\mid h)
\bigl(\log p_\theta(a\mid h)-\log q(a\mid h)\bigr)
\right],
\end{align}
because the extra score term has expectation zero. Hence
$-\nabla d_\theta(h)=\E[z_t r_t\mid h_t=h]$.
For $u>t$, $z_t$ is measurable at $h_u$, giving
\begin{equation}
\E[z_tr_u]=\E[z_t\E[r_u\mid h_u]]
=-\E[z_td_\theta(h_u)].
\end{equation}
These identities prove Eq.~\eqref{eq:sampled-credit}. They also show that the return-to-go estimator $\widehat G_{\mathrm{RTG}}=\sum_tz_tR_t$, with $R_t=\sum_{u=t}^{L}r_u$, satisfies $\E[\widehat G_{\mathrm{RTG}}]=-\nabla_\theta\mathcal K$.

\subsection{Alternative return-to-go proof of Proposition~\ref{prop:broadcast}}
\label{app:past-feedback}

Let $Z=\sum_tz_t$ and $P_t=\sum_{u<t}r_u$. Expanding the broadcast estimator gives
\begin{equation}
RZ=\sum_tz_t(P_t+R_t)
=\widehat G_{\mathrm{RTG}}+\sum_tz_tP_t.
\end{equation}
The past-feedback sum $P_t$ is determined by $h_t$, and $\E[z_t\mid h_t]=0$. Hence $\E[z_tP_t]=0$ and $\E[RZ]=\E[\widehat G_{\mathrm{RTG}}]=-\nabla_\theta\mathcal K$, as in Eq.~\eqref{eq:broadcast}. The direct sequence-likelihood proof appears in Section~\ref{sec:local}.

\subsection{Tail influence under exponential discounting}
\label{app:tail-influence}

We quantify the influence of later feedback on the trajectory score in Eq.~\eqref{eq:discounted-trajectory-score}. Fix a response length $L$, a prefix length $1\le H\le L$, and $0<\gamma<1$. Write $W_L=\sum_{t=1}^L\gamma^{t-1}$ and $a_{t,L}=\gamma^{t-1}/W_L$ for the normalized position weights. The total weight after $H$ is
\begin{equation}
\alpha_{\gamma,H,L}
=\sum_{t=H+1}^L a_{t,L}
=\frac{\gamma^H-\gamma^L}{1-\gamma^L}
\le \min\left\{\gamma^H,\frac{L-H}{L}\right\}.
\label{eq:tail-mass}
\end{equation}

\paragraph{Derivation of the tail weight.} Summing the geometric series gives the equality. The first bound follows from
\[
\alpha_{\gamma,H,L}
=\gamma^H\frac{1-\gamma^{L-H}}{1-\gamma^L}
\le\gamma^H.
\]
For the second bound, the weights $\gamma^{t-1}$ are nonincreasing, so their average over the last $L-H$ positions is no larger than their average over all $L$ positions. Dividing the tail sum by $W_L$ therefore gives $\alpha_{\gamma,H,L}\le(L-H)/L$. The case $H=L$ has zero tail weight.

\paragraph{Sensitivity to changes in later feedback.} Suppose two feedback sequences have the same length and agree through position $H$, with $|r'_t-r_t|\le\eta$ for $t>H$. Their trajectory scores satisfy
\begin{equation}
|S_L^{\gamma\prime}-S_L^\gamma|
=\left|\sum_{t=H+1}^L a_{t,L}(r'_t-r_t)\right|
\le\eta\alpha_{\gamma,H,L}.
\end{equation}
This follows directly from the triangle inequality and nonnegative weights. It requires no statistical noise model for the feedback.

The normalization also changes when a response is extended. If $|r_t|\le B$, compare the score $S_H^\gamma$ of a fixed prefix with the score $S_L^\gamma$ after appending a tail. For $H<L$, let $S_{H+1:L}^\gamma$ be the normalized weighted mean over that tail. Then
\begin{align}
S_L^\gamma
&=(1-\alpha_{\gamma,H,L})S_H^\gamma
+\alpha_{\gamma,H,L}S_{H+1:L}^\gamma,\\
|S_L^\gamma-S_H^\gamma|
&\le 2B\alpha_{\gamma,H,L}\le2B\gamma^H.
\end{align}
Thus the influence bound includes the change in the score denominator. At $\gamma=1$, the tail weight is $(L-H)/L$. These bounds quantify score sensitivity, not whether later feedback is useful. They do not establish an optimal discount or a guarantee of stable policy updates.

\subsection{Weighted objectives}
\label{app:weighted-objectives}

To relate position weighting to the teacher-matching objective, consider fixed nonnegative weights $w_1,\ldots,w_T$ that do not depend on the sampled response or its realized length. As in Section~\ref{sec:local}, the teacher is fixed and the student is sampled on-policy. We assume a common sequence space, teacher support wherever the student has positive probability, and interchangeability of differentiation and expectation. Terminated responses are padded to a fixed maximum length $T$ by a shared absorbing state with zero subsequent feedback. With $d_\theta(h_t)=\KL(p_\theta(\cdot\mid h_t)\|q(\cdot\mid h_t))$, define
\begin{equation}
R_w=\sum_{t=1}^{T}w_tr_t,
\qquad
\mathcal K_w=-\E_p[R_w]
=\sum_{t=1}^{T}w_t\E_p[d_\theta(h_t)].
\end{equation}
The last equality follows from $\E[r_t\mid h_t]=-d_\theta(h_t)$.

\paragraph{Decomposition into prefix objectives.} Let $K_n=\KL(p_{1:n}\|q_{1:n})$, where $p_{1:n}$ and $q_{1:n}$ are the distributions of the first $n$ positions, and set $K_0=0$. The KL chain rule gives
\[
K_n=\sum_{t=1}^{n}\E_p[d_\theta(h_t)],
\qquad
\E_p[d_\theta(h_t)]=K_t-K_{t-1}.
\]
Substitution and rearrangement of the finite sum, with $w_{T+1}=0$, yield
\begin{equation}
\begin{aligned}
\mathcal K_w
&=\sum_{t=1}^{T}w_t(K_t-K_{t-1})\\
&=\sum_{n=1}^{T}(w_n-w_{n+1})K_n.
\end{aligned}
\label{eq:prefix-kl-mixture}
\end{equation}
For nonincreasing $w$, all coefficients are nonnegative. A fixed prefix window $w_t=\mathbf{1}\{t\le H\}/H$ gives $\mathcal K_w=K_H/H$. Exponential weights $w_t=\gamma^{t-1}/W_T$ give coefficients $(1-\gamma)\gamma^{n-1}/W_T$ for $n<T$ and $\gamma^{T-1}/W_T$ for $n=T$. Thus these weighting schemes correspond to nonnegative combinations of prefix KL objectives.

\paragraph{Broadcast gradient for fixed weights.} Write $Z=\sum_{t=1}^{T}z_t$. Since the weights are fixed and $\nabla_\theta R_w=-\sum_t w_tz_t$,
\begin{equation}
-\nabla_\theta\mathcal K_w
=\E_p[R_w Z]+\E_p[\nabla_\theta R_w]
=\E_p[R_w Z],
\end{equation}
where the second term vanishes because $\E_p[z_t]=0$ at every position. This extends the raw broadcast identity to a fixed weighted objective. The practical score instead divides by the weight mass of each realized response length, and its update also uses group statistics and clipping. Those operations are outside this fixed-weight result, as discussed in Appendix~\ref{app:practical}.

\subsection{Algebraic interpretation of collective teacher pressure}
\label{app:pressure-calibration}

The scalar estimator in Eq.~\eqref{eq:gropd-estimator} has an equivalent group-level interpretation. Let $\mathbf S_g^\gamma=(S_1^\gamma,\ldots,S_G^\gamma)^\top$ and let $\mathbf 1$ be the all-ones vector. Define
\[
\mathbf C_g^\gamma=\mathbf S_g^\gamma-\bar S_g^\gamma\mathbf 1,
\qquad
\mathbf P_g^\gamma=\bar S_g^\gamma\mathbf 1,
\qquad
\mathbf S_g^\gamma=\mathbf C_g^\gamma+\mathbf P_g^\gamma.
\]
The relative response preference component $\mathbf C_g^\gamma$ records deviations from the group mean, while $\mathbf P_g^\gamma$ records collective teacher pressure. Both are constructed from the same teacher--student feedback. They are orthogonal under the Euclidean inner product because
\[
\langle\mathbf C_g^\gamma,\mathbf P_g^\gamma\rangle
=\bar S_g^\gamma\sum_{i=1}^{G}(S_i^\gamma-\bar S_g^\gamma)=0.
\]
For $\epsilon>0$, the advantage vector is therefore
\[
\mathbf A_g^{\gamma,\lambda}
=\frac{\mathbf S_g^\gamma-(1-\lambda)\bar S_g^\gamma\mathbf 1}{s_g^\gamma+\epsilon}
=\frac{\mathbf C_g^\gamma+\lambda\mathbf P_g^\gamma}{s_g^\gamma+\epsilon}.
\]
Its group mean and pairwise differences satisfy
\[
\frac1G\sum_{i=1}^{G}A_i^{\gamma,\lambda}
=\frac{\lambda\bar S_g^\gamma}{s_g^\gamma+\epsilon},
\qquad
A_i^{\gamma,\lambda}-A_j^{\gamma,\lambda}
=\frac{S_i^\gamma-S_j^\gamma}{s_g^\gamma+\epsilon}.
\]
Thus, for a fixed group of scores and its standard deviation, $\lambda$ changes only the shared offset and preserves all pairwise advantage differences and the response ordering. At $\lambda=0$, the shared offset vanishes and only the relative response preference component remains; this component still depends on teacher feedback. The shared offset is signed and may raise or lower the advantages together. These algebraic properties do not imply invariant PPO gradients or subsequent training trajectories. The decomposition interprets the scalar estimator without changing its computation or identifying separate semantic sources of guidance.

\section{Practical Estimator Boundaries}
\label{app:practical}

The expectation results above apply to raw returns and a single score-function update. The practical estimator divides the weighted trajectory score by its realized weight mass, scales advantages by a within-prompt group standard deviation, and uses clipping and repeated PPO updates. These data-dependent operations change the gradient estimator; the raw-return unbiasedness result does not transfer to the practical loss. The proofs motivate sequence-level credit assignment without asserting that the implemented update is the exact reverse-KL gradient.

\section{Experimental Details}
\label{app:details}

The five teacher--student pairings appear in Table~\ref{tab:main}. For Instruct models, think or nothink mode is part of the pairing and uses the corresponding chat template. Base models use their default reasoning protocol, which is distinct from the Instruct think/nothink mode switch. Both Instruct students use nothink mode; only the Instruct-teacher-to-Base-student pairings use a thinking Instruct teacher. Teacher and student log probabilities are evaluated on the same conditional history.

\paragraph{Parameter convention.} Throughout the paper, $\lambda$ is the fraction of collective teacher pressure retained. Archived run labels and training configurations use the complementary group-mean subtraction coefficient $c=1-\lambda$. All reported parameter values and figure legends use the retention convention; the underlying runs and measured results are unchanged.

\subsection{Training data}
\label{app:training-data}

We use the English configuration of \textsf{open-r1/DAPO-Math-17k-Processed}, pinned to revision 31dd3095, with 14,116 questions in its canonical training split. Each example retains its problem statement and reference answer and is converted into a system instruction followed by the user question. The instruction asks the model to reason step by step and place its final answer in a boxed expression. We apply no additional difficulty or correctness filtering. After chat-template rendering and tokenization, the training loader retains prompts with more than five and at most 4,096 tokens.

The fixed-response diagnostic draws questions from the same pool. Its candidate questions are deduplicated by normalized prompt text, and exact normalized matches to evaluation questions are excluded. A deterministic hash ordering with seed 20260910 selects 1,024 questions, whose alternating entries form two equal subsets. Figure~\ref{fig:sequence-credit-training}(b) uses the 512-question confirmation subset.

\subsection{Training configuration}
\label{app:training-config}

Table~\ref{tab:training-config} summarizes the GR-OPD training configuration. Each rollout batch contains 64 prompts with eight responses per prompt, yielding 512 student responses. Training uses the ROLL framework with Megatron for model optimization and teacher scoring, and vLLM for student generation.

The 512 responses are divided into eight global optimization minibatches of 64 responses each. The student uses eight data-parallel replicas, a configured per-device batch size of one, and eight gradient-accumulation steps. With one PPO epoch per rollout batch, each rollout step therefore produces eight optimizer updates.

\begin{table}[!htb]
\centering
\small
\caption{GR-OPD training configuration. Sampling settings refer to student rollouts during training.}
\label{tab:training-config}
\begin{tabular}{@{}ll@{}}
\toprule
Setting & Value \\
\midrule
Prompts per rollout batch & 64 \\
Responses per prompt, $G$ & 8 \\
Configured per-device training batch size & 1 \\
Gradient-accumulation steps & 8 \\
Student data-parallel replicas & 8 \\
Responses per global optimization minibatch & 64 \\
Optimizer updates per rollout step & 8 \\
Training rollout steps & 221 \\
Maximum prompt length & 4,096 tokens \\
Maximum response length & 28,672 tokens \\
Optimizer & Adam (Megatron implementation) \\
Learning rate & $5\times10^{-7}$ \\
Weight decay & $0.1$ \\
Learning-rate schedule & Linear warm-up followed by linear decay to zero \\
Warm-up & 20 optimizer steps \\
PPO epochs per rollout batch & 1 \\
PPO probability-ratio clipping, $\delta$ & $0.2$ \\
Sampling temperature / top-$p$ & $1.0$ / $1.0$ \\
Top-$k$ sampling & Disabled \\
Random seed & 42 \\
Model precision & bfloat16 \\
Reference hardware configuration & 8 NVIDIA H20 GPUs \\
\bottomrule
\end{tabular}
\end{table}

For GR-OPD, token-level teacher feedback is clipped to $[-5,5]$ before trajectory aggregation. After group calibration, the resulting teacher advantage is also clipped to $[-5,5]$ before the policy update. These two value-clipping operations are separate from the PPO probability-ratio clipping in Eq.~\eqref{eq:gropd-policy-objective}.

\paragraph{Terminal-token alignment.} Qwen3 Base and Instruct models use different end-of-sequence (EOS) conventions. For the Base-student/Instruct-teacher pairings, we map the student's EOS target (token ID 151643) to the teacher's EOS target (151645) when computing teacher log probabilities. This aligns the probability of termination under the two models. The mapping applies only to active response positions in the teacher-scoring copy; the sampled student response is preserved. Student generation also accepts token ID 151645 as an additional stopping token.

\paragraph{Model and tokenizer versions.} We use models from the \href{https://qwenlm.github.io/blog/qwen3/}{original Qwen3 release (April 2025)}. Table~\ref{tab:model-snapshots} maps the model names used in this paper to their official checkpoints. Each link pins a public revision containing the model weights and tokenizer files; we use each model's associated tokenizer. The models denoted Instruct correspond to the repositories without the \texttt{-Base} suffix.

\begin{table}[!htb]
\centering
\small
\setlength{\tabcolsep}{3pt}
\caption{Official checkpoints for the models and tokenizers. Repository links point to fixed public revisions on Hugging Face.}
\label{tab:model-snapshots}
\begin{tabular*}{\linewidth}{@{}l@{\extracolsep{\fill}}ll@{}}
\toprule
Model used in this paper & Official checkpoint & Public revision \\
\midrule
Qwen3-1.7B-Base & \href{https://huggingface.co/Qwen/Qwen3-1.7B-Base/tree/ea980cb0a6c2ae4b936e82123acc929f1cec04c1}{\texttt{Qwen/Qwen3-1.7B-Base}} & \texttt{ea980cb} \\
Qwen3-4B-Base & \href{https://huggingface.co/Qwen/Qwen3-4B-Base/tree/906bfd4b4dc7f14ee4320094d8b41684abff8539}{\texttt{Qwen/Qwen3-4B-Base}} & \texttt{906bfd4} \\
Qwen3-1.7B-Instruct & \href{https://huggingface.co/Qwen/Qwen3-1.7B/tree/70d244cc86ccca08cf5af4e1e306ecf908b1ad5e}{\texttt{Qwen/Qwen3-1.7B}} & \texttt{70d244c} \\
Qwen3-4B-Instruct & \href{https://huggingface.co/Qwen/Qwen3-4B/tree/1cfa9a7208912126459214e8b04321603b3df60c}{\texttt{Qwen/Qwen3-4B}} & \texttt{1cfa9a7} \\
Qwen3-32B-Instruct & \href{https://huggingface.co/Qwen/Qwen3-32B/tree/9216db5781bf21249d130ec9da846c4624c16137}{\texttt{Qwen/Qwen3-32B}} & \texttt{9216db5} \\
\bottomrule
\end{tabular*}
\end{table}

\paragraph{Software versions.} Our implementation is based on a modified ROLL 0.3.0 checkout with \texttt{mcore\_adapter} 0.10.0.dev0. The recorded training environment uses PyTorch 2.8.0, Transformers 4.57.1, Megatron-Core 0.16.1, and vLLM \texttt{0.11.1rc2.dev0+gc3a722fcb.d20251021}.

\subsection{Evaluation protocol}
\label{app:evaluation-protocol}

Table~\ref{tab:evaluation-protocol} lists the six evaluation suites and the number of sampled responses per question. For the main comparison, generation uses temperature $0.6$, top-$p=0.9$, no top-$k$ restriction, and a maximum response length of 28,672 tokens. Avg@K is the fraction of correct responses among the $K$ samples for each question, averaged over questions within a suite. Macro Avg@K is the unweighted mean of the six suite scores.

\begin{table}[htbp]
\centering
\small
\caption{Evaluation question counts and samples per question. The six suites contain 1,042 questions in total.}
\label{tab:evaluation-protocol}
\begin{tabular}{@{}lrr@{}}
\toprule
Benchmark & Questions & $K$ \\
\midrule
AIME 2024 & 30 & 32 \\
AIME 2025 & 30 & 32 \\
AMC 2023 & 40 & 32 \\
MATH-500 & 500 & 8 \\
Minerva Math & 272 & 8 \\
OlympiadBench (math subset) & 170 & 8 \\
\bottomrule
\end{tabular}
\end{table}

Correctness is determined by the ROLL mathematical-answer verifier using Math-Verify. After validating the thinking-tag structure, the verifier examines the text after the closing thinking tag, falling back to the final paragraph when no closing tag is present. It first extracts the last boxed expression; if no complete boxed answer is available, it applies the mathematical-expression parser to that answer text. The extracted result is checked for mathematical equivalence to the reference answer. Failed extraction, parsing errors, and verification timeouts receive a score of zero. Responses that reach the generation limit remain in the evaluation and are judged from the generated text using the same procedure.

\subsection{Teacher and student base capabilities}

Table~\ref{tab:base-capabilities} provides pre-OPD model-level reference scores. Its rows have different generation caps and are not used as matched method comparisons. Mean response length is computed by re-tokenizing the saved response text with the corresponding tokenizer, without adding special tokens; prompt tokens are excluded.

\begin{table}[h]
\centering
\small
\setlength{\tabcolsep}{3.6pt}
\caption{Pre-OPD reference capability. Macro Avg@K and mean response length are averaged equally over the six suites. Mean response length is measured in tokens.}
\label{tab:base-capabilities}
\begin{tabular}{@{}llcrr@{}}
\toprule
Model & Mode & Role & Macro Avg@K & Mean tokens \\
\midrule
Qwen3-1.7B-Base & default & Student & 23.099 & 4,166.4 \\
Qwen3-4B-Base & default & Teacher & 31.244 & 3,245.5 \\
Qwen3-1.7B-Instruct & nothink & Student & 32.995 & 1,662.1 \\
Qwen3-4B-Instruct & nothink & Teacher & 42.970 & 1,937.5 \\
Qwen3-4B-Instruct & think & Teacher & 74.574 & 10,138.4 \\
Qwen3-32B-Instruct & think & Teacher & 78.954 & 9,221.9 \\
Qwen3-32B-Instruct & nothink & Teacher & 44.163 & 2,152.6 \\
\bottomrule
\end{tabular}
\end{table}

\subsection{Fixed-response horizon diagnostic}
\label{app:fixed-response-diagnostic}

Figure~\ref{fig:sequence-credit-training}(b) measures how teacher scores from different prefix lengths distinguish correct from incorrect responses on a fixed rollout bank. For each of 512 prompts, we score 32 responses over their first $\min(H,L_i)$ tokens, averaging teacher--student log-ratios clipped to $[-5,5]$. Responses are organized into groups of eight. Let $S_{gi}^H$ be the score of response $i$ in group $g$, let $C_{gi}\in\{0,1\}$ indicate correctness of its complete answer, and let $\mathcal G_x$ be the complete groups for prompt $x$. The plotted statistic is
\begin{equation}
J_H(H)=\frac{1}{512}\sum_{x=1}^{512}
\frac{1}{|\mathcal G_x|}\sum_{g\in\mathcal G_x}
\frac{\operatorname{Cov}_{i\in g}(S_{gi}^H,C_{gi})}
{\operatorname{Std}_{i\in g}(S_{gi}^H)}.
\label{eq:jh-diagnostic}
\end{equation}
Covariance and standard deviation use their population definitions. A group contributes zero if its score standard deviation is at most $10^{-6}$ or if all its correctness labels agree; such groups remain in the average. We first average groups within each prompt and then average prompts equally. Positive values mean that correct responses receive higher standardized teacher scores on average. This fixed-response test compares scoring horizons on the same responses; it does not compare policies learned by training with different horizons.

\subsection{Training diagnostics for feedback horizon}
\label{app:horizon-training-diagnostics}

\begin{figure}[H]
\centering
\includegraphics[width=\linewidth]{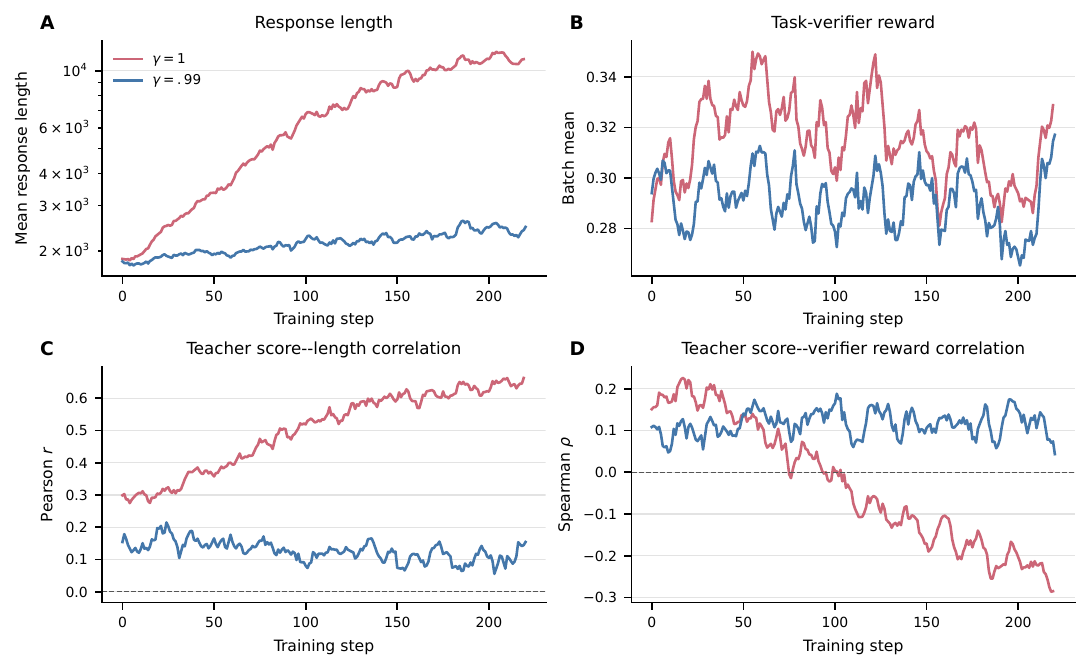}
\caption{Training dynamics for $\gamma=1$ and $\gamma=.99$ on \setIIThirtyTwo{}. (A) Mean response length in tokens (logarithmic axis); (B) mean task-verifier reward; (C) Pearson correlation between trajectory score and response length; and (D) Spearman correlation between trajectory score and verifier reward. Correlations pool responses within each rollout batch. Curves use centered moving averages as specified above.}
\label{fig:horizon-training-dynamics}
\end{figure}

Figure~\ref{fig:horizon-training-dynamics} compares $\gamma=1$ and $\gamma=.99$ on \setIIThirtyTwo{}. Both runs use full group centering ($\lambda=0$). At each rollout step, let $L_i$ be the number of active response tokens, $v_i$ the task-verifier reward, and $S_i^\gamma$ the trajectory score before group calibration. The score aggregates token-level teacher feedback clipped to $[-5,5]$ using Eq.~\eqref{eq:discounted-trajectory-score}. Panels (A) and (B) report the batch means of $L_i$ and $v_i$. Panels (C) and (D) compute
\begin{equation}
r_{SL}=\operatorname{Corr}_{i\in\mathcal B}(S_i^\gamma,L_i),
\qquad
\rho_{Sv}=\operatorname{Corr}_{i\in\mathcal B}
\bigl(\operatorname{rank}(S_i^\gamma),\operatorname{rank}(v_i)\bigr),
\end{equation}
where $\operatorname{Corr}$ is Pearson correlation and $\mathcal B$ contains the responses in the rollout batch. Spearman correlation uses average ranks for ties; the implementation returns zero when the correlation denominator is at or below the numerical tolerance. These correlations pool responses across prompts, without within-prompt centering, and differ from the grouped correctness statistic $J_H(H)$. For visualization, we use centered moving averages over seven rollout steps for (A), (C), and (D), and eleven for (B), using available observations at the endpoints.

Under $\gamma=1$, response length grows and the trajectory score becomes increasingly associated with length, while its rank association with verifier reward turns negative. With $\gamma=.99$, length grows more slowly, score--length correlation remains weaker, and score--verifier correlation remains positive. This contrast suggests that a trajectory score can increasingly favor length without continuing to favor correct answers. The correlations describe the evolving batches and do not establish that length causes the change or isolate the contribution of particular continuations.

\subsection{Pressure magnitude diagnostics}
\label{app:pressure-diagnostic}

Table~\ref{tab:shared-score-component} reports batch-level magnitude proxies at $\gamma=.99$ for completed 221-step runs. For each training batch, we divide the absolute batch mean trajectory score by the batch mean within-prompt group standard deviation plus $10^{-8}$. The normalized proxy is the median of this ratio over training steps; the retained proxy multiplies it by $\lambda$. This ratio of batch summaries is not an average of per-prompt normalized pressure magnitudes and does not measure the signed shared component for an individual prompt.

\paragraph{Pressure magnitude varies across pairings.} The normalized proxy ranges from $3.16$ to $3.61$ in the two matched Instruct pairings, compared with $6.08$ and $7.23$ in the two cross-pattern pairings. Thus, a larger shared score offset relative to within-prompt variation accompanies pattern mismatch in these runs. Within each matched pairing, the normalized proxy varies little across the reported retention settings: $3.16$--$3.21$ for \setIIFour{} and $3.56$--$3.61$ for \setIIThirtyTwo{}. Most of the difference in the retained proxy therefore comes from the retention fraction. The same fraction can nevertheless retain different magnitudes across pairings: at $\lambda=.25$, the proxies are $0.79$ and $0.90$, respectively.

\paragraph{Magnitude must be interpreted with training behavior.} For \setIIFour{}, $\lambda=.5$ retains a proxy magnitude of $1.60$ and eventually leads to reward decline, whereas $\lambda=.25$ and $.2$ retain $0.79$ and $0.63$ and support sustained improvement (Figure~\ref{fig:training-controls-matrix}). The completed cross-pattern runs retain no shared component at $\lambda=0$, while their relative response preference term still uses teacher feedback. Together, these observations support calibrating collective teacher pressure to the pairing. They do not establish a universal magnitude threshold or show that a large shared component is intrinsically harmful: the proxy summarizes the magnitude of collective teacher pressure, while its teaching value also depends on compatibility with the student's current behavior.

\end{document}